\documentclass{article} % For LaTeX2e
\usepackage{iclr2024_conference,times}

\usepackage{hyperref}
\usepackage{booktabs}	%% For formal tables:
\usepackage{subcaption} %% For complex figures with subfigures/subcaptions
\usepackage{listings}
\usepackage{xspace}
\usepackage[utf8]{inputenc}
\usepackage[T1]{fontenc}
\usepackage{xcolor,colortbl}
\usepackage[normalem]{ulem}
\usepackage{footnote}
\usepackage{pdfpages}
\usepackage{multicol}
\usepackage{multirow}
\usepackage{graphics}
\usepackage{todonotes}
\usepackage{wrapfig}
\usepackage{comment}
\usepackage{tikz}
\usepackage{url}
\usepackage{tablefootnote}
\usepackage[framemethod=tikz]{mdframed}
\usepackage{subcaption}

\usetikzlibrary{shapes,arrows,positioning,automata}

\newcommand{\rebuttal}[1]{#1}

\newcommand{\smalltextsc}[1]{\textsc{\small #1}}
\newcommand{\scripttextsc}[1]{\textsc{\scriptsize #1}}

\newcommand{\greenbg}[1]{\colorbox{green!25}{#1}}
\newcommand{\yellowbg}[1]{\colorbox{yellow!25}{#1}}

\newcommand{\llm}{\smalltextsc{LLM}\xspace}
\newcommand{\llms}{\smalltextsc{LLMs}\xspace}

\newcommand{\vllm}{\smalltextsc{vLLM}\xspace}
\newcommand{\dquality}{\textit{data quality}\xspace}
\newcommand{\lmdt}{\smalltextsc{LMDT}\xspace}

\newcommand{\cllamaB}[1]{\smalltextsc{CodeLLaMa-#1B}\xspace}
\newcommand{\cl}{\smalltextsc{CL-7B}\xspace}
\newcommand{\codedavinci}{\smalltextsc{Code-Davinci-002}\xspace}
\newcommand{\gptturbo}{\smalltextsc{GPT-3.5-turbo}\xspace}
\newcommand{\gptfour}{\smalltextsc{GPT-4}\xspace}
\newcommand{\gptfourturbo}{\smalltextsc{GPT-4-Turbo}\xspace}
\newcommand{\alphacode}{\smalltextsc{AlphaCode}\xspace}
\newcommand{\alphacodenine}{\smalltextsc{AlphaCode-9B}\xspace}
\newcommand{\alphacodefour}{\smalltextsc{AlphaCode-41B}\xspace}

\newcommand{\apps}{\smalltextsc{APPS}\xspace}
\newcommand{\appsintro}{\smalltextsc{APPS-Introductory}\xspace}
\newcommand{\appsinter}{\smalltextsc{APPS-Interview}\xspace}
\newcommand{\appscompete}{\smalltextsc{APPS-Competition}\xspace}
\newcommand{\contests}{\smalltextsc{Code-Contests}\xspace}
\newcommand{\contestsplanfilter}{\smalltextsc{Code-Contests-Plan}\xspace}

\newcommand{\passmetric}[1]{\smalltextsc{Pass@}#1\xspace}

\newcommand{\origD}{$\mathcal{D}_{original}$\xspace}
\newcommand{\renameD}{$\mathcal{D}_{rename}$\xspace}
\newcommand{\modularD}{$\mathcal{D}_{modular}$\xspace}

\newcommand{\planD}{$\mathcal{D}_{planning}$\xspace}
\newcommand{\plangtD}{$\mathcal{D}_{plan}^{GT}$\xspace}
\newcommand{\distillD}{$\mathcal{D}_{distill}$\xspace}
\newcommand{\modularDfour}{$\mathcal{D}_{modular}^{4}$\xspace}
\newcommand{\modularDthreefive}{$\mathcal{D}_{modular}^{3.5}$\xspace}

\definecolor{codegreen}{rgb}{0,0.6,0}
\definecolor{codegray}{rgb}{0.5,0.5,0.5}
\definecolor{codepurple}{rgb}{0.58,0,0.82}
\definecolor{backcolour}{rgb}{0.95,0.95,0.92}
\definecolor{mauve}{rgb}{0.58,0,0.82}

\lstdefinestyle{pythonstyle}{
    language=Python,
    backgroundcolor=\color{backcolour},
    commentstyle=\color{codegreen},
    keywordstyle=\color{magenta},
    numbers=none,
    numberstyle=\tiny\color{codegray},
    stringstyle=\color{codepurple},
    basicstyle=\tiny\ttfamily,
    breakatwhitespace=false,
    breaklines=true,
    captionpos=b,
    keepspaces=true,
    numbersep=5pt,
    showspaces=false,
    showstringspaces=false,
    showtabs=false,
    tabsize=2,
    keywords={for, while, and, print, dfs, join, read_input, apply_operations, print_output, main, calculate_distances, process_queries, continue, bfs, build_graph, gcd, dp, binary_search, check_if_infinite, calculate_sum_points, group_books_by_genre, sort_books_by_price, calculate_purchase_prices, calculate_max_purchase_price, get_factors, get_largest_divisor, ncr, is_good_number, count_excellent_numbers, generate_palindromes, find_closest_palindrome, find_divisors, calculate_min_sum, find_root, merge_trees, find_possible_book_locations, calculate_possible_locations, find_max_distance, count_possible_locations, calculate_beauty, count_sequences, find_dividers, calculate_max_colors, solution, pop, append, print_grid, remove_white_rows, remove_white_columns, read_grid, bisect,
    find_permutation, calculate_permutation_odd, calculate_permutation_even,count_operations, calculate_change, update_heights},
}

\lstdefinestyle{planstyle}{
    language={},
    backgroundcolor=\color{backcolour},
    keywordstyle=\color{magenta},
    numbers=none,
    numberstyle=\tiny\color{codegray},
    basicstyle=\tiny,
    captionpos=b,
    breakindent=0pt,
    showtabs=false,
    tabsize=2,
    keywords={dfs, join, read_input, apply_operations, print_output, main, calculate_distances, process_queries, continue, bfs, build_graph, gcd, dp, binary_search, check_if_infinite, calculate_sum_points},
}

\newcommand{\code}[1][]{\lstinline[basicstyle=\small\ttfamily,#1]}

\usepackage{amsmath,amsfonts,bm}

\def\eqref#1{equation~\ref{#1}}
\def\1{\bm{1}}

\def\rvd{{\mathbf{d}}}

\def\rvy{{\mathbf{y}}}

\DeclareMathAlphabet{\mathsfit}{\encodingdefault}{\sfdefault}{m}{sl}
\SetMathAlphabet{\mathsfit}{bold}{\encodingdefault}{\sfdefault}{bx}{n}

\def\gD{{\mathcal{D}}}

\def\gI{{\mathcal{I}}}

\def\gM{{\mathcal{M}}}

\def\gO{{\mathcal{O}}}

\newcommand{\showcomments}{yes}
\newcommand\weilin[1]{\ifthenelse{\equal{\showcomments}{yes}}{{\color{cyan} Wei-Lin: #1}}{\ignorespaces}}

\title{LLM-Assisted Code Cleaning For Training Accurate Code Generators}

\author{Naman Jain, Tianjun Zhang, Wei-Lin Chiang, Joseph E. Gonzalez, Koushik Sen \& Ion Stoica\\
University of California, Berkeley\\
\texttt{\{naman\_jain,tianjunz,weichiang,jegonzal,ksen,istoica\}@berkeley.edu}
}

\iclrfinalcopy % Uncomment for camera-ready version, but NOT for submission.
\begin{document}
\maketitle

\begin{abstract}
    %There has been a lot of interest in building stronger code generation systems 
Natural language to code generation is an important application area of \llms{} and has received wide attention from the community. 
The majority of relevant studies have exclusively concentrated on increasing the quantity and functional correctness of training sets while disregarding other stylistic elements of programs.
More recently, data quality has garnered a lot of interest and multiple works have showcased its importance for improving performance.
In this work, we investigate data quality for code and find that making the code more structured and readable leads to improved code generation performance of the system.
 % functionally equivalent programs 
% simply improving the structure and readability of the programs leads to better performance performance.
% quality for code. 
% We identify the importance of structured and readable code as well as high-level planning which is necessary for programming. 
We build a novel data-cleaning pipeline that uses these principles to transform existing programs by 1.) renaming variables, 2.) modularizing and decomposing complex code into smaller helper sub-functions, and 3.) inserting natural-language based plans via \llm{} based transformations.
% More specifically, our pipeline re-writes existing programs by improving the variable names, decomposing complex program functionality into smaller sub-functions and finally inserting high-level \textit{planning} annotations using natural language. 
We evaluate our approach on two challenging algorithmic code generation benchmarks and find that fine-tuning \cllamaB{7} on our transformed modularized programs improves the performance by up to \textbf{30\%} compared to fine-tuning on the original dataset. 
Additionally, we demonstrate improved performance from using a smaller amount of higher-quality data, finding that a model fine-tuned on the entire original dataset is outperformed by a model trained on 15\% of our cleaned dataset.
Even in comparison to closed-source models, our models outperform the much larger \alphacode{} models~\citep{li2022competition}.

% Data quality is a central factor contributing to the effectiveness of large language models (\llms), improving model performance during both the pre-training and fine-tuning phases
% This work addresses the challenge of understanding and elevating data quality in the context of code generation.
% Focusing on improving the high-level reasoning and low-level programming, we identify variable renaming (making variable names contextually meaningful), function decomposition (splitting the code into meaningful sub-programs), and planning (providing high-level planning in the form of natural language descriptions) as key aspects of data quality.
% We propose a novel approach, Language Model Based Data Transformations (\lmdt{} for brevity), to automatically construct such high-quality datasets by using \llms{} to transform a given dataset.
% We apply \lmdt{} to transform training sets from two challenging algorithmic code generation benchmarks, \apps{} and \contests{} in accordance with the identified properties.
% Our findings reveal that models finetuned on our transformed data improve code generation performance by up to \confirm{30\%} compared to those finetuned on the original data sets.
% \weilin{one thing we could highlight here is the use of ``oracle equivalence checker'' to ensure data quality. I do think it's is more principled against other methods such as wizardlm-evol.}

% % data quality is important
% % recent works
\end{abstract}
\vspace{-0.4cm}
\section{Introduction}
\label{sec:intro}
%  first what people doing wrong

% high level
% we are first to clean code
% show llms do it quite well
% surprising effect of even simple transformations 

% approach and pipeline in more detail

% datasets, results and findings
% widely used datasets 
% functional correctness metric
% hardness of dataset
% performance improvement 

% lessons -- learned

Natural language to code generation has witnessed considerable advances in recent years with the advent of large language models (\llms{} for brevity). 
These advances primarily arise from training on large web-scale data and are measured based on the functional correctness of the programs. 
Thus, other aspects like readability, structuring, and styling and how they affect training and data quality are largely ignored by these works.
%\confirm{[need a connecting sentence along the lines of -- do not consider the kind of data used for training]}
On the flip side, many recent works have demonstrated the effectiveness of training on higher quality data during both pre-training~\citep{li2023textbooks2} and fine-tuning~\citep{zhou2023lima,cao2023instruction} phases.
Even within the code-generation domain, ~\cite{gunasekar2023textbooks} demonstrated the benefits of training on a ``textbook'' quality dataset, generated synthetically using the \gptturbo{} model~\citep{ouyang2022training}. 
However, these works do not provide an understanding of the factors that actually improve the data quality.

In this work, we show that using programs following good programming practices and allowing for more readability leads to improved code generation performance compared to using programs that do not follow these practices.
We use these insights to build a novel automated code data-cleaning pipeline that transforms programs while maintaining functional correctness using input-output examples. 
In contrast to prior works that curate \textit{high quality} datasets by directly generating \textit{new} data using \llms{}, here we translate existing datasets into their \textit{parallel cleaned} versions while identifying attributes that actually improve data quality.
%Specifically, we focus on attributes such as lexical naming and control-flow logic of the programs
%transform programs to rename variables to be contextually relevant, decompose complex functionality into smaller helper sub-functions, and insert high-level planning annotations to orchestrate the program logic.
%Specifically, we rename variable names, decompose complex program functionality into smaller helpful sub-functions and finally insert high-level natural language planning annotations. 
%we transform and clean existing datasets constructing their \textit{parallel} versions. 

We use \llms{} to perform the transformations used in our data-cleaning approach.
We demonstrate that instruction-tuned models can take a user-identified attribute of data quality as a natural language instruction and perform the transformation accurately. 
Our approach leverages the disparity in difficulty between generating a solution and editing an existing one.
Therefore, it is particularly effective in domains where the existing model struggles to generate a correct solution but can effectively edit a given solution.
We perform our data-cleaning transformations in three iterations: 1) renaming variables 2) modularizing complex code into subfunctions, and 3) adding planning annotations. 

\begin{figure}[!t]
    \centering
    \includegraphics[width=\textwidth]{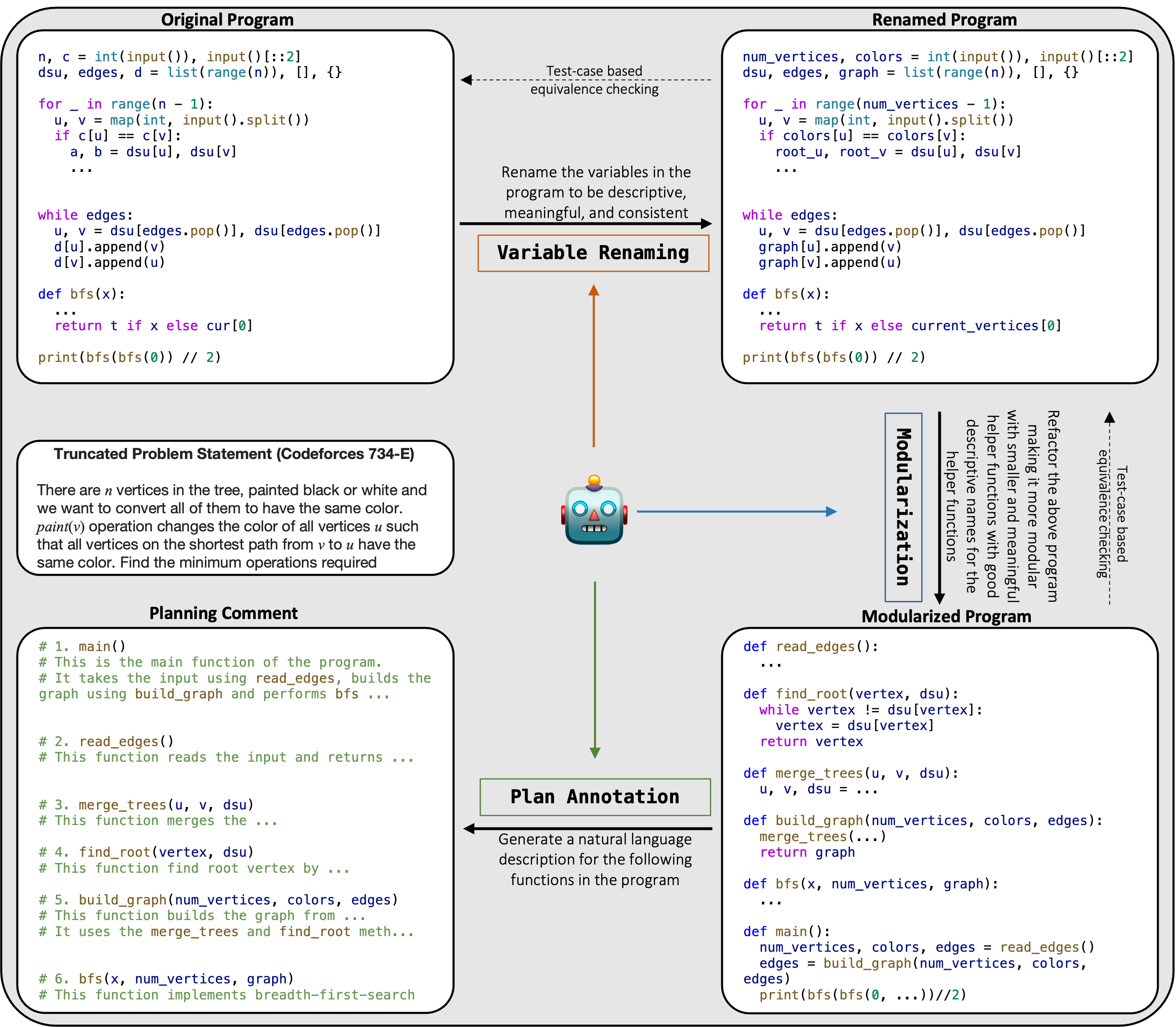}
    \vspace{-0.4cm}
    \caption{\textbf{The overview of our code cleaning approach}. We apply instruction-tuned \llms{} to transform existing datasets by providing natural language prompts and use input-output examples to maintain function equivalence between original and transformed programs. Our cleaning approach works in three steps. The top-left figure depicts the original program from the dataset. This program first undergoes variable renaming (top-right figure). Next, the renamed program is decomposed into constituent sub-functions and converted into a modularized program (bottom-right figure). Finally, we generate a natural-language \textit{plan} from the modularized program by summarizing the functions in a top-down manner (bottom-left figure). \rebuttal{This plan is prepended to the program as a comment.} The middle-left figure presents the truncated problem statement}
    \label{fig:intro:lmdt_diag}
    \vspace{-0.6cm}
\end{figure}

Figure~\ref{fig:intro:lmdt_diag} provides an overview of our approach. 
Notice that the variable renaming step at the top adjusts the variable names to be contextually relevant (e.g. \code{a} to \code{root_u} and \code{d} to \code{graph}). 
The modularization step (depicted on the right) identifies and decomposes the original program into several smaller subfunctions such as \code{find_root}, \code{merge_trees}, \code{build_graph}, etc. It then implements these subroutines and assembles the modular program. 
Finally, our planning step (depicted at the bottom) constructs a plan by summarizing  functions in a top-down fashion (starting from the \code{main}). 

% \begin{figure}[!t]
%     \centering
%     \usetikzlibrary{automata,shapes,shapes.geometric,arrows,fit,calc}
%     \tikzset{->,elliptic state/.style={draw,ellipse}}
%     \begin{tikzpicture}[shorten >=1pt, scale = 0.8, transform shape]
%         \node[elliptic state, align=center] (lmdt) {\lmdt{}};
%         \node[elliptic state, align=center, left of=lmdt, yshift=0.9cm, xshift=-2.5cm] (description) {Transformation\\Description};
%         \node[elliptic state, align=center, left of=lmdt, yshift=-0.9cm, xshift=-2.5cm] (program) {Program};
%         \node[elliptic state, align=center, right=of lmdt] (transformed) {Transformed\\Program};

%         \draw (program) -- (lmdt);
%         \draw (description) -- (lmdt);
%         \draw (lmdt) -- (transformed);
%     \end{tikzpicture}
%     \caption{An overview of our \lmdt{} approach applied to transforming programs. The user provides a transformation description as a natural language instruction such as renaming variables, modularize programs and our approach applies the transformation to the program. Optionally, the user can provide an oracle checker which ensure similarity of transformed output with the original input, for example, here, ensuring functional equivalence between the original and transformed programs.}
%     \label{fig:intro:lmdt_diag}
% \end{figure}

We evaluate our approach in a niche, yet challenging, domain of algorithmic code generation.
The goal is to generate a program for a given problem statement.
The task is challenging because it requires both high-level algorithmic reasoning and low-level coding and is evaluated using a strict functional correctness metric.
%Figure~\ref{fig:intro:lmdtex1} provides an example of successful transformations applied to a program.
%\confirm{Notice that the renaming step adjusts the variable names to be contextually relevant (e.g. \code{is_infinite} instead of \code{b}). The modularization step identifies various smaller helper sub-functions in the program, such as \code{read_input}, \code{calculate_distances} and \code{process_queries}, and the planning step adds a natural language description of the program.}
%\input{figures/introduction/lmdtex1}
We use two well-known algorithmic code generation benchmarks, namely \apps~\citep{hendrycksapps2021} and \contests~\citep{li2022competition}.
We transform the corresponding programs in the training sets and obtain \textit{parallel} datasets from our cleaning approach.
Additionally, we utilize input-output examples to maintain functional equivalence between the original and transformed programs.
We qualitatively analyze the generated dataset and find that it uses smaller helper sub-functions, each often implementing a standard algorithm or key program functionality, 
%Even in programs that correspond to different problems, we find shared helper functionalities implements -- such as graph traversals, shortest-path calculations, binary search, and more! 
and provide more in depth findings in Section~\ref{subsec:exp:data-analysis}.
We further assess the impact of the transformed datasets on the performance on our downstream code generation task. 
We fine-tune \cllamaB{7} model on the various collected datasets. 
Our findings reveal that the model fine-tuned on our modularized dataset outperforms the model fine-tuned on the functionally equivalent original dataset by up to \textbf{30\%}.
Beyond, performance improvement, we also demonstrate that improving data quality improves the data efficiency. In particular, a model fine-tuned on the entire original dataset is outperformed by a model trained on just 15\% of our cleaned dataset.
%\confirm{Similarly, zero-shot code execution performance on data samples from cleaned datasets improves by AAA.}

We next study improving planning in a supervised learning setup similar to prior works~\citep{fu2023specializing,li-etal-2023-symbolic}.
While we observe limited improvements in planning, we disentangle planning vs coding capabilities and find that our fine-tuned model is capable of using gold-annotated plans, extracted from the ground-truth solutions to accurately generate solutions for the complex programs. 
This highlights planning for complex problems remaining a key bottleneck that does not seem to improve by merely increasing training datasets.
%Our findings reveal that while fine-tuning plans does not improve the \textit{reasoning} only minimally improves the reasoning capability of the models, the models can accurately use ground truth natural language plans to solve problems much better. 
%Similarly, variable renaming, a sub-step in our cleaning process also improves the performance over the original dataset but is superseded by the modularized dataset.
Finally, in comparison to existing baselines, our fine-tuned models outperform the larger \alphacode{}~\citep{li2022competition} models. 
%Finally, the programs generated by our approach

% In summary, our contributions are as follows

% \begin{itemize}
%     \item We propose a simple yet effective pipeline for automatic curation of high-quality datasets. This is achieved by transforming and enhancing existing datasets based on specified data cleaning criteria.
%     \item We apply our approach to the domain of algorithmic code generation by transforming programs to contain contextually relevant variable names, smaller helper sub-functions, and high-level planning explanations.
%     \item Finally, we show that finetuning on our refined datasets improves downstream code-generation capability by upto \confirm{30\%} while also improving the data efficiency.
% \end{itemize}

\section{Methodology}
\label{sec:methods}
In this section, we present our general data transformation approach and then instantiate it for performing code data cleaning.

\subsection{Transformations for data cleaning}
Given a dataset $\gD$ consisting of $N$ instances $\rvd_i$, such that, $\gD = \{\rvd_i\}_{i=1}^N$.
To achieve a desired data cleaning specification, the user additionally provides a data-cleaning instruction $\gI$, which highlights an attribute that needs to be modified. 
Optionally, we also use an oracle equivalence checker ($\gO$) which ensures that the transformed data instance $\tilde{\rvd_i}$ is consistent with the original input based on some desired metric.
For example, we can use edit-distance or functional equivalence based on input-output examples as our oracle checker. 
%\rebuttal{Note that our method can accommodate a learned ``oracle'' as well  necessary for free-form generations.}

We use a pre-trained language model (denoted by $\gM$) to generate the transformed instance ($\tilde{\rvd_i}$) by prompting the model with the transformation instruction ($\gI$) and the original answer ($\rvy$).
We can perform either zero-shot or few-shot prompting for performing the data cleaning operation. 
Finally, we extract the instance $\tilde{\rvd_i}$ generated by $\gM$, and apply our oracle equivalence checker ($\gO$) to ensure consistency with the original data. 
If $\gO(\tilde{\rvd_i}, \rvd_i)=0$, i.e., the oracle reports a failure, we reject the generated output and retry the example within a sampling budget.
%In case of a rejection, we allow a fixed number of retry attempts.

While our transformation approach does not provide any guarantees about the quality of the performed transformation and relies on \llms{}, 
we empirically observe that instruction-tuned \llms{}  can perform various unstructured data cleaning steps quite effectively. 
%Additionally, using constraints can ensure transformation \llms{}
%effective in generating high-quality outputs. 
We provide a detailed analysis of the generated outputs for our algorithmic code generation setting in Section~\ref{subsec:exp:data-analysis}. 
Finally, in accordance with existing literature on prompting \llms{}, we found that using simple and precise, low-level instructions improves the performance and accuracy of the models in performing the operations. 
Thus, for complex data cleaning operations (refactoring), we find improvements by breaking it down and performing multiple operations iteratively (renaming followed by modularization).
%in accordance with existing prompting literature, we observe that providing precise, low level, and iterative instructions fare better than providing single high-level instruction.

\subsection{Code Data-Cleaning}
We apply our transformations-based data cleaning approach to programming data. 
Coding requires both -- low-level programming and high-level reasoning or planning skills.
Therefore, we propose a three-step cleaning pipeline that improves the readability and program structuring targeting the low-level coding skills and inserts natural-language based plans data targeting the high-level reasoning skills.
Our steps are detailed below.
%We posit that readability and program structuring improve the coding performance of \llms{} and learning from large amounts of diverse \textit{plans} improv 
% We identify readability and program structuring as essential 

\begin{enumerate}
    \item \textbf{Rename variables.} 
    This step renames the variables in the program, making them descriptive and easier to follow. 
    Figure~\ref{fig:intro:lmdt_diag} top provides an example of this transformation.
    \item \textbf{Modularize functions.} 
    Problem decomposition has been identified as a key approach for improving the reasoning capabilities of models~\citep{zhou2022least,wang2023plan}. 
    We identify program decompositions and transform the program by extracting their functionality into smaller helper functions. 
    Figure~\ref{fig:intro:lmdt_diag} right provides an example of this transformation.
    \item \textbf{Plan annotations.} 
    This step summarizes the helper functions in the already modularized program and prepends it to the programs in the form of a natural language plan. 
    These natural language descriptions are analogous to prompting approaches that are used for solving reasoning problems like chain-of-thought prompting~\citep{wei2022chain}, parsel~\citep{zelikman2023parsel}, etc. 
    Figure~\ref{fig:intro:lmdt_diag} bottom provides an example of this transformation.
\end{enumerate}

Additionally, while performing these transformations, we use the test cases provided in the dataset to construct our oracle equivalence checker ($\gO$). It ensures that our transformed programs maintain functional equivalence to the original program. 

\section{Experimental Setup}
\label{sec:exp}
In this section, we detail our experimental setup and implementation. Section~\ref{subsec:setup:data} outlines the benchmarks and metrics used for the algorithmic code generation task, while Sections ~\ref{subsec:setup:lmdt} and~\ref{subsec:setup:experiments} delve into the specifics of our code cleaning approach and fine-tuning experiments respectively.

\subsection{Benchmarks}
\label{subsec:setup:data}
We use two standard algorithmic code generation benchmarks,  \apps{} and \contests{}.
The benchmarks provide a collection of problem statements described in natural language and corresponding test cases.
The goal is to generate a program that successfully solves the problem.
The evaluation is performed using a strict functional correctness metric.
%We present details about the respective datasets and metrics below.

\textbf{\apps~\citep{hendrycksapps2021}.}
This benchmark includes 10,000 problems, evenly split between training and test sets.
It is sourced from multiple open-access competitive programming websites.
It is further divided into \appsintro{}, \appsinter{}, and \appscompete{} subsets based on problem difficulty.
In this study, we only consider problems sourced from a subset of the competition websites based on the number of test cases provided.
%In this work, we only consider problems sourced from websites provided sufficient number of training test cases.

\textbf{\contests~\citep{li2022competition}.}
This benchmark includes 13,328 problems in the training set and 165 problems in the test set.
We only use a subset of the training split that includes \verb|python| solutions satisfying the provided test cases.
Additionally, since the training set provides over a hundred solutions per problem, we perform LSH based near-deduplication on the solutions and limit them to a maximum of 25 solutions per problem.

Table~\ref{tab:setup:datasets} and Appendix~\ref{sec:app:setup} provide further details about our final datasets.

\begin{table}[!t]
    \centering
    \resizebox{1\columnwidth}{!}{%
        \begin{tabular}{|r|l|c|c|c|c|}
            \hline
                            & split & \appsintro & \appsinter & \appscompete & \contests \\
            \hline
            \multirow{2}{*}{Problems count}     & train & 42         & 1247       & 361          & 7132      \\
            & test  & 702        & 2699       & 309          & 165       \\
            \multirow{2}{*}{Tests count} & train & 1          & 1          & 9            & 200       \\
             & test  & 10         & 19         & 39           & 200       \\
            Solutions count    & train & 736        & 18394      & 5060         & 98582     \\
            \hline
        \end{tabular}
    }
    \vspace{-0.2cm}
    \caption{Details about the number of problems, the median number of test cases per problem, and the number of solutions in the \apps{} and \contests{} datasets.}
    \label{tab:setup:datasets}
    \vspace{-0.6cm}
\end{table}

\textbf{Metrics.}
We assess the code generation performance of the models using the \passmetric{$K$} metric~\citep{kulal2019spoc,chen2021evaluating}, which evaluates the functional correctness of generated programs.
For each problem, we generate $N$ solutions (where $N \geq 2K$) and compute the expected number of scenarios in which the problem is solved at least once when sub-selecting a random sample of $K$ solutions.
We vary $K$ in $\{1, 10, 25\}$ for \apps{} dataset and $\{1, 10, 100\}$ for the \contests{} benchmark. We present more details about sampling hyperparameters in Appendix~\ref{sec:app:setup}.

\subsection{Data Transformations}
\label{subsec:setup:lmdt}
We apply our data transformation approach on the \apps{} and \contests{} datasets.
Unless specified otherwise, we use \gptturbo{} as our default language model $\gM$ to perform the transformations and use a default temperature $0.3$.
In case of failure, we retry up to 5 iterations.
We obtain three \textit{parallel} datasets at the end of our cleaning process, one for each of renaming, modularization, and planning \rebuttal{(note that the transformations are applied sequentially)}. Table~\ref{tab:setup:transformed-datasets} provides a summary of the generated datasets along with the instructions used to generate them. We provide complete details about the transformations in Appendix~\ref{sec:app:transformimpl}.

We also simulate a simple direct synthetic data generation approach somewhat similar to~\cite{gunasekar2023textbooks}. 
Specifically, we generate solutions for the training problems using the \gptturbo{} model.
We use in-context learning with the two-shot prompt examples selected from our \modularD{} dataset.
To ensure diverse solutions, we use three distinct few-shot examples and generate eight solutions for every prompt at a temperature of $0.5$.
\rebuttal{Additionally, we filter the solutions for correctness based on the ground truth test cases provided in the dataset to ensure we are not training on incorrect programs. 
Since it resembles a distillation-like setup, we refer to this dataset as \distillD{}.}

%To simulate a direct data-generation baseline, we also collect a dataset by generating solutions for the training problems using the \gptturbo{} model. We use a two-shot prompt, by selecting the in-context learning examples from \modularD{}. We refer to this dataset as \distillD{}.

\begin{table}
    \centering
    \resizebox{1\columnwidth}{!}{%
        \begin{tabular}{|l|l|l|p{12cm}|}
            \hline
            \textbf{Dataset} & \textbf{Notation}                            & \textbf
            {Applied On}     & \textbf{Transformation Instruction} ($\gI$)                                                                                                                                                                            \\
            \hline
            Base             & \origD                                       & -         & -                                                                                                                                                            \\

            Rename           & \renameD                                     & \origD    & \textit{Rename the variables in the program to be descriptive, meaningful, and consistent}                                                                   \\

            Modularize       & \modularD                                    & \renameD  & \textit{Refactor the above program making it more modular with smaller and meaningful helper functions with good descriptive names for the helper functions} \\

            Plan             & \planD                                       & \modularD & \textit{Generate a natural language description for the following functions in the program}                                                                  \\
            \hline
        \end{tabular}
    }
    \vspace{-0.2cm}
    \caption{Transformed datasets generated by our code cleaning approach. For each transformation, we have provided the corresponding notation, the transformation instruction used to perform the cleaning step and the dataset the transformation was applied on.}
    \label{tab:setup:transformed-datasets}
    \vspace{-0.6cm}
\end{table}

\subsection{Experiment Details}
\label{subsec:setup:experiments}
To evaluate the \textit{quality} of the transformed datasets, we measure how they impact the test benchmark accuracy.
We study both in-context learning and fine-tuning using examples from our datasets.
%Additionally, we also construct a distillation or direct solution generation baseline and compare our transformed dataset against it.

\textbf{Models.}
We use the \cllamaB{7} model~\citep{roziere2023codellama} in all our experiments (referred as \cl{} ahead).
We use the model checkpoint from huggingface\footnote{\url{https://huggingface.co/codellama/CodeLlama-7b-hf}}
and perform batched inference through \vllm{}~\citep{kwon2023efficient}, necessary for computing the \passmetric{$K$} metric.
We also present the numbers from \codedavinci{} and \gptturbo{} whenever available.

\textbf{In-context learning.}
We select two question-answer pairs from the \origD{} and \modularD{} training sets as our in-context learning example.
For a fair comparison between the two evaluations, we use the same problem and corresponding solutions from the two datasets as examples.
The examples are combined with appropriate delimiters and the model is then prompted with a new problem.
Note that these in-context learning examples increase the sequence length by over 2,000 tokens and considerably slow the inference.

\textbf{Fine-Tuning.}
We perform full fine-tuning over the base \cl{} model on the different datasets.
%(i.e. \origD{}, \renameD{}, \modularD{}, \planD{}, and \distillD{})
We train the models for two epochs on the \apps{} dataset and one epoch on the \contests{} dataset using a $5e^{-5}$ learning rate and an effective batch size of 256 on 4 A6000 GPUs.

\section{Experimental Results}
\label{sec:results}
We present our experimental results in this section. Section~\ref{subsec:exp:data-analysis} first provides a qualitative overview of the transformed programs and Section~\ref{subsec:exp:main-results} presents the main code generation results.

\subsection{Analysis of the transformed programs}
\label{subsec:exp:data-analysis}
\begin{table}[!t]
    \centering
    \resizebox{0.85\columnwidth}{!}{
        \begin{tabular}{lrrrrrrr}
            \hline
                                   &                                &                         &                                &                 &                         &                         &                         \\[-0.33cm]

                                   & \multicolumn{3}{c}{\appsintro} &                         & \multicolumn{3}{c}{\appsinter}

            \\ \cline{2-4}  \cline{6-8}
                                   &                                &                         &                                &                 &                         &                         &                         \\[-0.33cm]

                                   & \passmetric{1}                 & \passmetric{10}
                                   & \passmetric{25}                &                         & \passmetric{1}                 & \passmetric{10} &
            \passmetric{25}                                                                                                                                                                                                    \\
            \hline
                                   &                                &                         &                                &                 &                         &                         &                         \\[-0.33cm]
            \textbf{In-context Learning}
                                   &                                &
                                   &                                &                         &                                &                 &
            \\
            \hline
                                   &                                &                         &                                &                 &                         &                         &                         \\[-0.33cm]
            \cl{} + \origD{}
                                   & 14.2                           & 29.2
                                   & 38.4                           &                         & 1.8                            & 7.3             & 10.4
            \\
            \cl{} + \modularD{}
                                   & 17.5                           & 30.1
                                   & 39.7                           &                         & 2.2                            & 8.6             & 12.3
            \\
                                   & \greenbg{+3.3}                 & \greenbg{+0.9}
                                   & \greenbg{+1.3}                 &                         & \greenbg{+0.4}                 & \greenbg{+1.3}  & \greenbg{+1.9}
            \\
            \hline
                                   &                                &                         &                                &                 &                         &                         &                         \\[-0.33cm]
            \textbf{Fine-tuning}
                                   &                                &
                                   &                                &                         &                                &                 &
            \\
            \hline
                                   &                                &                         &                                &                 &                         &                         &                         \\[-0.33cm]
            \cl{} + \origD{}
                                   & 18.7                           & 34.4
                                   & 40.2                           &                         & 3.4                            & 9.7             & 13.6
            \\
            \cl{} + \modularD{}
                                   & \textbf{22.7}                  & \textbf{36.9}
                                   & 42.6                           &                         & \textbf{4.2}                   & \textbf{11.0}   & \textbf{15.0}
            \\
                                   & \greenbg{\textbf{+4.0}}        & \greenbg{\textbf{+2.5}} & \greenbg{\textbf{+2.4}}        &                 & \greenbg{\textbf{+0.8}} & \greenbg{\textbf{+1.3}} & \greenbg{\textbf{+1.4}} \\
            \cl{} + \planD{}
                                   & 22.1                           & \textbf{37.1}
                                   & \textbf{43.8}                  &                         & 3.7                            & 10.5            & 14.8
            \\
            \cl{} + \renameD{}
                                   & 19.2                           & 36.6
                                   & 42.9                           &                         & 4.0                            & 10.7            & 14.6
            \\
            % Distillation-1
            %                 & 19.3                              & 33.7
            %                 & 39.1                              &                                 &                 &
            %                 &                                   &                                 &                       \\
            \cl{} + \distillD{}
                                   & 21.1                           & 35.3
                                   & 40.5                           &                         & \rebuttal{4.1}                 & 10.8            & 14.5
            \\
            \hline
                                   &                                &                         &                                &                 &                         &                         &                         \\[-0.33cm]
            \textbf{Closed models}
                                   &                                &
                                   &                                &                         &                                &                 &
            \\
            \hline
                                   &                                &                         &                                &                 &                         &                         &                         \\[-0.33cm]
            \codedavinci\tablefootnote{Model generations were obtained from
            ~\cite{chen2022codet}} & 22.1                           & 50.2
                                   & 58.7                           &                         & 4.1                            & 16.8            & 23.8
            \\
            \hline
        \end{tabular}}
    \vspace{-0.2cm}
    \caption{\textbf{Results on \apps{} dataset.} We use the \cllamaB{7} model (referred to as \cl{}) under in-context learning and fine-tuning. We use samples from the original and our transformed datasets and find that our cleaned datasets improve the performance of the model by over \textbf{20\%}. The green highlighted numbers depict the improvements obtained from using \modularD{} (over \origD{}). Similarly, using \renameD{} and \planD{} also provide improvements, usually lesser than using \modularD{}.}
    \label{tab:apps_results}
    \vspace{-0.5cm}
\end{table}

\textbf{Data statistics.}
For the \contests{} dataset, out of 98,582 programs extracted from the original dataset (\origD), we can successfully transform 92,675 (94.0\%) into our modularized dataset (\modularD{}).
We obtain similar success rates for the \apps{} dataset (details deferred to the appendix).
On the contrary, the distilled dataset (\distillD{}), which is constructed by generating solutions directly using \gptturbo{} only finds a correct solution for about 50\% of the problems.

\textbf{Analysis of the transformed programs.}
We find that our transformation approach decomposes the original programs by inserting three new functions on a median ($\scriptstyle \sim$2.6 functions on average).%, thus ensuring a granular control flow in the program.
To get a better understanding of the decomposition, we cluster the functions using their function names and signatures.
We find that these helper functions often implement key program logic, standard algorithms, and utilities like handling inputs, outputs, and orchestrating the main function.
Interestingly, we also find that the helper functions are often reused across problems, with small variations in implementations.
For example, the top five most frequent helper functions, \code{dfs}, \code{build_graph}, \code{gcd}, \code{dp}, and \code{binary_search} occur in about 3-8\% of the problems.
Additionally, we qualitatively analyze a hundred random samples from \origD{} and \modularD{} datasets to determine the quality of performed transformations.
Figures~\ref{fig:appendix:example-1} to~\ref{fig:appendix:example-10} in the appendix provide examples of such transformations.
We find that most of the transformations are meaningful. 
They improve the readability of the programs and also find suitable decomposition for the program logic encoded in the control flow (see Figures~\ref{fig:appendix:example-1},~\ref{fig:appendix:example-2},~\ref{fig:appendix:example-3},~\ref{fig:appendix:example-11} as examples).
%The control flow of the resulting transformed programs additionally provides high-level algorithmic logic and is useful for planning (see Figure~\ref{fig:appendix:example-1} as an example).
However, in some cases, the generated helper functions can have improper names (\code{calculate_max_colors} in Figure~\ref{fig:appendix:example-10}) or complex implementations copied directly from the original program (\code{count_sequences} in Figure~\ref{fig:appendix:example-9}).
Additionally, for simpler programs (Figure~\ref{fig:appendix:example-8}), the entire program functionality can be implemented in a single function and the \textit{decomposition} does not provide any extra information. \rebuttal{Finally, we use \gptfour{} as judge~\citep{zheng2023judging} evaluation to quantitatively assess the transformations in regards to their meaningfulness and about the consistency of original and transformed programs. Appendix~\ref{app:sec:gpt4_eval} presents the comprehensive setup. We find that over 99\% of the transformations are regarded as helpful of which only 3-5\% of examples are judged as \textit{can do better}. Similarly, 99.4\% of the transformed programs are judged as consistent with the original programs. More detailed evaluation results in Table~\ref{app:tab:gpt4_judge}.}

Unlike, generated code, we cannot constrain or check the generated natural language plans.
Thus, we find that sometimes the plans can be imprecise and vary in detail. 
While using a stronger pretrained model like \gptfour{} could alleviate some of these issues, we believe this will be a good avenue for applying something analogous to process supervision~\citep{lightman2023let}.

\subsection{Main Results}
\label{subsec:exp:main-results}
Tables~\ref{tab:apps_results} and~\ref{tab:contests_results} provide our primary results on \apps{} and \contests{} datasets respectively. We defer the results for the \appscompete{} subset to Appendix~\ref{sec:app:results} and highlight our findings below.

\subsubsection{Effect of modularization}
We find that our data-cleaning approach improves the performance of the model on both \apps{} and \contests{} datasets in both in-context learning and fine-tuning settings. 
\begin{table}

    \centering
    \begin{minipage}[!t]{0.49\textwidth}
        \resizebox{\columnwidth}{!}{
            \begin{tabular}[!t]{lrrr}
    %\hline
    \firsthline
                                                                               &                                 &                         &                         \\[-0.35cm]
                                                                               & \multicolumn{3}{c}{\contests{}}                                                     \\[0.03cm]
    \cline{2-4}
                                                                               &                                 &                         &                         \\[-0.35cm]
                                                                               & \passmetric{10}                 & \passmetric{25}         & \passmetric{100}        \\[0.03cm]

    \hline
                                                                               &                                 &                         &                         \\[-0.35cm]
    \textbf{In-context Learning}                                               &                                 &                         &                         \\[0.03cm]
    \hline
                                                                               &                                 &                         &                         \\[-0.35cm]
    \cl{} + \origD{}                                                           & 5.1                             & 6.5                     & 7.2                     \\[0.03cm]
    \cl{} + \modularD{}                                                        & 4.9                             & 6.6                     & 9.3                     \\[0.03cm]
                                                                               & \yellowbg{-0.2}                 & \yellowbg{+0.1}         & \greenbg{+2.1}          \\[0.03cm]
    \hline
                                                                               &                                 &                         &                         \\[-0.35cm]
    \textbf{Fine-tuning}                                                       &                                 &                         &                         \\[0.03cm]
    \hline
    \cl{} + \origD{}                                                           & 5                               & 6.4                     & 10.9                    \\[0.03cm]
    \cl{} + \modularD{}                                                        & \textbf{6.1}                    & \textbf{8.3}            & 12.4                    \\[0.03cm]
                                                                               & \greenbg{\textbf{+1.1}}         & \greenbg{\textbf{+1.9}} & \greenbg{\textbf{+1.5}} \\
    \cl{} + \planD{}                                                           & 5.3                             & 7.0                     & 10.8                    \\[0.03cm]
    \cl{} + \renameD{}                                                         & 4.7                             & 6.3                     & 10.5                    \\[0.03cm]

    \hline
                                                                               &                                 &                         &                         \\[-0.35cm]
    \textbf{Closed models}                                                     &                                 &                         &                         \\[0.03cm]
    \hline
                                                                               &                                 &                         &                         \\[-0.35cm]

    \alphacodenine\tablefootnote{Result sourced from~\cite{li2022competition}} & 5.0                             & 7.0                     & 10.0                    \\[0.03cm]
    \alphacodefour\footnotemark[3]                                             & 5.0                             & 7.0                     & 10.0                    \\[0.03cm]
    \codedavinci\tablefootnote{Result sourced from~\cite{zhang2023algo}}       & 3.0                             & -                       & 7.5                     \\[0.03cm]
    \gptturbo{}\tablefootnote{Result sourced from~\cite{li2023think}}          & -                               & -                       & 18.2                    \\[0.03cm]
    \text{  } + \textsc{Brainstorm}\footnotemark[5]                            & -                               & -                       & 29.3                    \\
    \hline
\end{tabular}

        }
        \vspace{-0.2cm}
        \caption{\textbf{Result on the \contests{} dataset.} Similar to findings on the \apps{} dataset, we find that our data cleaning approach generally improves the performance with modularization working particularly well while planning and renaming providing marginal to no improvements.}
        \label{tab:contests_results}

    \end{minipage}%
    \hfil%
    \begin{minipage}[!t]{0.47\textwidth}
        \resizebox{\columnwidth}{!}{
            \begin{tabular}[!t]{lrrr}
    \hline
    & & & \\[-0.3cm]
& \multicolumn{3}{c}{\contestsplanfilter{}}
\\[0.05cm]
\cline{2-4}
    & & & \\[-0.3cm]
                      & \passmetric{10} & \passmetric{25} & \passmetric{100} \\[0.05cm]
    \hline
    & & & \\[-0.3cm]
    \cl{} + \origD{}    & 6.5             & 9.5             & 15.0             \\[0.05cm]
    \cl{} + \modularD{} & 8.8             & 11.8            & 17.8             \\[0.05cm]
    \cl{} + \planD{}    & 6.9             & 10.5            & 15.4             \\[0.05cm]
    \cl{} + \plangtD{}  & 17.9            & 22.3            & 28.1             \\[0.05cm]
     & \greenbg{\textbf{+9.1}}            & \greenbg{\textbf{+10.5}}            & \greenbg{\textbf{+11.3}}             \\
    \hline
\end{tabular}

% \begin{table}
%     \centering
%     \resizebox{0.5\columnwidth}{!}{%
%         \begin{tabular}{|c|c|c|c|}
%             \hline
%                                      & \passmetric{10} & \passmetric{25} & \passmetric{100} \\
%             \hline
%             \cl{}-\origD{}    & 6.5             & 9.5             & 15.0             \\
%             \cl{}-\modularD{} & 8.8             & 11.8            & 17.8             \\
%             \cl{}-\planD{}    & 6.9             & 10.5            & 15.4             \\
%             \cl{}-\plangtD{}  & 17.9            & 22.3            & 28.1             \\
%             \hline
%         \end{tabular}}
%     \caption{Results on a subset of \contestsplanfilter{} dataset using ground truth plans}
%     \label{tab:results:gtplanresults}
% \end{table}
        }
        \vspace{-0.2cm}
        \caption{\textbf{Effect of using ground-truth plans.} We disentangle the high-level reasoning vs coding capabilities by extracting ground-truth plans from solutions corresponding to the test problems. We find significant improvement in the performance on the \contestsplanfilter{} dataset, indicating that the model trained on the \planD{} dataset while incapable of building correct plans, can follow such plans accurately.}
        \label{tab:results:gtplanresults}

    \end{minipage}
    \vspace{-0.6cm}

\end{table}

% \begin{table}[!t]
%     \begin{subtable}[t]{.49\linewidth}
%         \centering
%         \resizebox{\columnwidth}{!}{
%             \input{tables/contests-results}
%         }
%     \end{subtable}%
%     \hfil%
%     \begin{subtable}[!t]{.47\linewidth}
%         \centering
%         \resizebox{\columnwidth}{!}{
%             \input{tables/contests-gtplan-results}
%         }
%         \caption{\textbf{Effect of using ground-truth plans.} We disentangle the high-level reasoning vs coding capabilities by extracting ground-truth plans from solutions corresponding to the test problems. We find significant improvement in the performance on the \contestsplanfilter{} dataset, indicating that the model trained on the \planD{} dataset while incapable of building correct plans, can follow such plans accurately}
%         \label{tab:results:gtplanresults}
%     \end{subtable}
% \end{table}

% \begin{table}
%     \centering
%     \resizebox{0.55\columnwidth}{!}{
%         \input{tables/contests-results}
%     }
%     \caption{Results on \contests{} dataset}
%     \label{tab:contests_results}
% \end{table}

% \begin{table}
%     \centering
%     \resizebox{0.45\columnwidth}{!}{
%         \input{tables/contests-gtplan-results}
%     }
%     \caption{Effect of using ground-truth plans on the \contestsplanfilter{} dataset}
%     \label{tab:contests_results}
% \end{table}

\textbf{In-context Learning.}
We first evaluate the performance of the model when provided with \textit{parallel} two-shot in-context learning examples from \origD{} and \modularD{} datasets each. 
We find that the \passmetric{1} improves from 14.2 to 17.5 (a 23\% relative improvement) on the \appsintro{} dataset and \passmetric{100} improves from 7.2 to 9.3 (a 29\% relative improvement) on the \contests{} dataset.
These results indicate that more readablity and better-structured coding is helpful to the model in solving more problems.

\textbf{Fine-tuning.}
Next, we fine-tune the model on the \origD{} and \modularD{} datasets and again find strong performance improvements from our transformation approach.
Specifically, on the \appsintro{} dataset, the \passmetric{1} improves from 18.7 to 22.7 (a 23\% relative improvement). Similarly, the \contests{} dataset \passmetric{25} metric improves from 6.4 to 8.4 (30\% relative improvement).
These results cement our above findings about the effect of cleaning the data.

Interestingly, we also note that fine-tuning only provides modest improvements over the in-context learning performance.
We hypothesize that this is due to the challenging nature of our task.
\footnote{Note that the in-context examples add over 2,000 tokens to the prefix and lead to much slower decoding}
%that program decomposition and modularization allow the model to solve more complex problems more easily.
%Interestingly, we observe that finetuning only provides modest improvements over the in-context learning baselines.
%In fact, on the \contests{} dataset, finetuning on \origD{} does not improve the \passmetric{10}.

\subsubsection{Effect of planning annotations}
\label{subsec:exp:planning}
Prior work has demonstrated considerable successes in improving reasoning in \llms{} ~\citep{yue2023mammoth,magister2022teaching,fu2023specializing} by performing supervised learning on natural language \textit{reasoning} or \textit{planning steps}. 
We perform similar experiment, fine-tuning the model on \planD{} dataset consisting of plans generated by our approach on top of \modularD{}.
We find that planning only provides a modest improvement over the \modularD{} dataset (\passmetric{25} improved from 42.6 to 43.9 on the \appsintro{} dataset) or often no improvements at all. 
%These results are in sharp contrast to the literature on improving mathematical reasoning where training on chain-of-thought reasoning has been shown to be particularly effective.
%We hypothesize that this is an artifact of the difficult nature of our task and the diverse \textit{reasoning} required for solving these problems. We leave a thorough comparison of these tasks for future work.

Upon inspection of the generated solutions, we find that often the generated plans are imprecise or incorrect, highlighting that planning still remains a bottleneck. 
To disentangle the high-level planning from the coding component, we analyze the performance of the model when provided with ground-truth plans on the \contests{} dataset.
We extract these ground-truth plans by applying our data transformation approach on the test set (similar to how \planD{} training set was created).
Table~\ref{tab:results:gtplanresults} provides results on this subset of 109 problems from the \contests{} dataset for which we were able to extract the ground truth plans (since some problems don't have a valid \verb|python| solutions).
While our model trained on the \planD{} dataset is incapable of synthesizing new plans, it can follow the generated plans correctly.
All metrics improve significantly, e.g. \passmetric{100} improving from 17.8 to 28.1, well over the performance of \gptturbo{}, a much larger model! 

\rebuttal{
Our mixed results raise critical questions for future work on improving planning in \llms{}. In particular, poor performance might be attributed to any \textit{imprecision} in automatically generated plans. Future data curation techniques that filter or augment this imprecision would be valueable. 
Alternatively, the supervised learning paradigm followed in this work might be insufficient for models to generalize \textit{planning} in complex domains. Future work can explore alternative learning algorithms, possibly over our modularization approach which naturally decomposes programs.
}
\vspace{-0.32cm}

\begin{figure}
    \centering
    \begin{minipage}[!t]{0.45\textwidth}
        \begin{mdframed}[leftmargin=0pt,rightmargin=0pt,innerrightmargin=3pt,innerleftmargin=3pt,innertopmargin=0.4pt,innerbottommargin=0.4pt,roundcorner=5pt,backgroundcolor=backcolour]
    \input{figures/results/generated-prog}
    \end{mdframed}
    \vspace{-0.3cm}
    \caption{Example of a program generated by our model trained on the \modularD{} dataset. It solves the problem by using helper functions acting on rows and columns.} \label{fig:results:casestudy}
    \end{minipage}\hfil%
    \input{figures/results/datasize_plot}
    \vspace{-0.3cm}
\end{figure}

\subsubsection{Ablations}
\label{subsec:res:ablation}

\paragraph{Effect of data size.} Beyond improving the quality of the resulting model, data quality is also attributed to improving the data efficiency.
We evaluate this aspect by fine-tuning our model on different fractions of \origD{} and \modularD{} datasets and find similar results.
Figure~\ref{fig:results:datasize} presents the performance of the model as a function of training set size.
As shown in the figure, training on just 15\% of \modularD{} dataset achieves similar \passmetric{1} as fine-tuning on the entire \origD{}.
%less than \textbf{15\%} of the \modularD{} dataset provides similar performance as training on the entire \origD{} dataset.

\textbf{Effect of renaming.}
We use variable renaming as an intermediate step in our cleaning process.
We evaluate the performance of the model fine-tuned only on the \renameD{} dataset and find that renaming provides some performance improvements when compared to fine-tuning on \origD{} dataset. For example, \passmetric{1} improved from 17.2 to 19.1 on \appsintro{}.
However, renaming still performs worse in comparison to fine-tuning on the \modularD{}. 
This highlights that beyond just readable code, functional decomposition is also a key aspect of improving our performance.

\textbf{Cleaning Transformations vs Distillation.}
We compare our transformation approach with a direct distillation baseline where we directly generate solutions using \gptturbo{}, referred to as the \distillD{} dataset\footnote{Note that we generate these solutions using in-context examples from the \modularD{} dataset}.
\rebuttal{This corresponds to various \llm{} instruction or fine-tuning approaches~\citep{xu2023wizardlm,li-etal-2023-symbolic} providing a strong baseline for data cleaning.}
On the \appsintro{} dataset, we find that fine-tuning on the \modularD{} dataset achieves better performance compared to the \distillD{} dataset demonstrating the advantage of cleaning over the generation baseline. 

\rebuttal{
\textbf{Choice of transformation model.} 
To evaluate how the choice of transformation model affects performance, we use the \gptfourturbo{} model to transform on a subset of the training set (detailed setup in Appendix~\ref{subsec:app:model_ablation}). 
\gptfourturbo{}, a stronger model, performs the transformations successfully and the resulting model trained on this version of the modularized dataset achieves even higher accuracy.
%compared to models trained on \textit{parallel} set from \gptturbo{}. 
For instance, \passmetric{10} improves from 33.0 when using \modularD{} constructed with \gptturbo{} to 34.3 when using the \modularD{} constructed with \gptfourturbo{} (full results in Table~\ref{tab:app:model_ablation}).
}
\vspace{-0.4cm}
\subsubsection{Comparison to Other Baselines}
Beyond \cl{}, fine-tuned models outperform strong baselines like \alphacode{} on the \contests{} dataset but still lag behind larger \codedavinci{} and \gptturbo{} models.

\subsubsection{Case study of generated modularized program}
Figure~\ref{fig:results:casestudy} provides an example of a program correctly generated by a model fine-tuned on our \modularD{} dataset. 
The problem requires removing rows and columns containing cells with certain attributes (i.e., if the cell is white)
The modularized solution correctly identifies the steps required to solve the problem and implements them as separate helper functions, providing readable code.
%The modularized solution correctly identifies the steps required to solve and implements as separate functions allowing more readable and explainable code.

\section{Related Work}
\label{sec:related}
%Data quality has been receiving increasingly more attention in the \llm{} literature, both in terms of improving the model performance and also for reducing the data requirements.
%-- both for improving the performance and reducing the data requirements.

\textbf{Instruction tuning.} Instruction tuning refers to the process of finetuning a base pretrained \llm{} to perform general-purpose tasks and follow instructions. Recent works, ~\cite{zhou2023lima,cao2023instruction,chen2023maybe} have demonstrated that a small high-quality instruction corpus is sufficient for achieving good instruction tuning performance.
Here, we perform task-specific fine-tuning of \llms{} and observe similar performance improvements.

\textbf{Synthetic data for \llms{}.}\ \ \ 
Recent works have explored using synthetic datasets for general-purpose or task-specific finetuning of \llms{}.
These approaches work by generating synthetic datasets from a strong \llm{} (like \gptturbo or \gptfour) using a set of existing tasks~\citep{alpaca, vicuna2023} or generating new tasks using self-instruct~\citep{wang2022self} or evol-instruct~\citep{xu2023wizardlm} approaches.
This has been also applied for task-specific finetuning -- in common-sense reasoning~\citep{west-etal-2022-symbolic}, text-summarization~\citep{sclar2022referee}, mathematical reasoning ~\citep{luo2023wizardmath,yue2023mammoth}, tool use~\citep{patil2023gorilla}, coding~\citep{luo2023wizardcoder}, and general-purpose reasoning~\cite{li-etal-2023-symbolic, zelikman2022star}.

More specifically, ~\cite{yue2023mammoth} curates diverse corpus of mathematics problems with chain-of-thought or program-of-thought~\citep{chen2022program} annotations for mathematical reasoning analogous to our plans.
~\cite{gunasekar2023textbooks} proposed pre-training models on programming ``textbooks'' generated synthetically from \gptturbo{}.
~\cite{haluptzok2023language} similarly generates programming puzzles and corresponding solutions from language models. 
Our work also studies curating synthetic data for code-generation space. 
However, instead of directly generating data using \llms{}, we identify good programming patterns and clean existing datasets using them. 
%Unlike these works, instead of directly generating data using \llms{}, we identify data characteristics and use \llms{} to transform existing data to collect higher-quality data.

\textbf{Algorithmic Code Generation.}
Code generation is a broad domain and is covered in Appendix~\ref{sec:app:related}. We only discuss pertinent algorithmic code generation works here. 
~\cite{hendrycksapps2021} released the \apps{} dataset while ~\cite{li2022competition} released the \contests{} dataset with the \alphacode{} models.
~\cite{zhang2023planning} proposed a lookahead-search-based decoding algorithm for improving \textit{reasoning} in \llms{} and is orthogonal to our work. 
~\cite{chen2022codet,zhang2023algo} proposed  \scripttextsc{CodeT} and \scripttextsc{ALGO}, that use generated tests (using \llm{} or brute-force solution) to re-rank the generated solutions.
~\cite{zelikman2023parsel} proposed the \textsc{Parsel} approach which used the \codedavinci{} model to first generate a plan in their high-level problem-specification language and then generate a program using it. 
~\cite{li2023explaining} also study disentangling the planning and code generation capabilities for closed source \llms{}, similar to our experiments on open models. Finally, recent work~\cite{le2023codechain} proposed a prompting based approach for modular code-generation.

\section{Discussion and Conclusion}
\label{sec:discuss}
Traditionally, data quality has been linked to functional correctness, ignoring the rich stylistic aspects differing across programs.
In this work, we demonstrate that these aspects like readability, and program structuring actually impact the performance of the trained model on downstream tasks and thus also contribute to \dquality{}.
\rebuttal{
    Next, we proposed a novel data-cleaning pipeline demonstrating that \llms{}
    can be used for transforming existing datasets to improve their quality based on user-instructions and oracle equivalence checker.
    While our evaluations focused on the algorithmic code generation task, we believe that this approach would also be useful for other domains for improving data quality as well. In particular, even in the absence of symbolic checkers (like test cases), we believe that there is an opportunity to use learned ``oracles'' for ensuring consistency and quality in other domains akin to how used in ~\cite{sclar2022referee}. Finally, beyond improving algorithmic code generation, we believe our modularization approach can be beneficial for general software engineering use cases (test generation, debugging, verification) where modularity is beneficial.
}
%In this work, we proposed \lmdt{}, a novel approach to collect high-quality datasets by transforming existing datasets based on user-provided instructions.
%We identified improving lexical and control-flow structure of code improves low-level coding and high-level planning skills.
%Experimental results demonstrate that our approach improves the performance of \llms{} on algorithmic code generation task.

\paragraph{Acknowledgement} This work was supported in part by NSF grants CCF-1900968, CCF-1908870 and by SKY Lab industrial sponsors and affiliates Astronomer, Google, IBM, Intel, Lacework, Microsoft, Mohamed Bin Zayed University of Artificial Intelligence, Nexla, Samsung SDS, Uber, and VMware. Any opinions, findings, conclusions, or recommendations in this paper are solely those of the authors and do not necessarily reflect the position of the sponsors.
Additionally, we thank Alex Gu, Manish Shetty, and anonymous reviewers for helpful discussion and feedback on the paper.

\bibliographystyle{iclr2024_conference}
\bibliography{main}

\appendix
\newpage
\section{Experimental Setup}
\label{sec:app:setup}
\paragraph{\apps{} benchmark.}
Since some of the problems in the \apps{} dataset are sourced from websites that provide insufficient or absent test cases, we filter the problems from those platforms.
Specifically, we only retain problems from the \verb|codeforces|, \verb|codechef|, and \verb|atcoder| competition websites.
This also removes the disparity/domain-shift between the training and test splits which has been observed as an issue in the \apps{} dataset in prior works (Section 4.1 in \cite{li2023think}).
While we considerably reduced the size of our training set, our test set is still quite close to the test set containing around 3800 problems instead of the default 5000.

\paragraph{\contests{} benchmark.}
The original \contests{} benchmark consists of 13,328 problems in the training dataset. We restrict the dataset to only problems with valid \verb|python| solutions that pass the test cases. \rebuttal{
Next, since the original dataset provides over 100 solutions per problem, we perform minhash-based deduplication on the solutions (hash size=64, num bands=60, band size=5) from gaoya\footnote{\url{https://github.com/serega/gaoya}} and retain a maximum of 25 solutions per problem. This results in about 7k problems in the training set spanning about 98.5k solutions. We do not perform any filtering on the test set.

Additionally, we note that some of the provided solutions in both \apps{} and \contests{} datasets do not pass the test cases.
These cases are sometimes caused by incorrect programs, corresponding to solutions in the wrong programming language.
However, more often this is caused by problems in these datasets supporting multiple correct solutions (for instance solutions can return a list of elements in any order).
The provided test cases only check for a single correct solution and thus result in many solutions failing the test cases.
We retain such samples for the smaller \apps{} dataset and use original \textit{correct} programs for matching output behavior instead of provided outputs.
}

\paragraph{Metrics}
We use the \passmetric{$K$} to perform our evaluations. We perform nucleus sampling using \vllm{} with $p=0.95$. We outline the default sampling configurations used for computing the metrics
\begin{enumerate}
    \item \passmetric{1} - We use a sampling budget ($N$) = 10 and temperature = 0.1.
    \item \passmetric{10} - We use a sampling budget ($N$) = 50 and temperature = 0.6.
    \item \passmetric{25} - We use a sampling budget ($N$) = 50 and temperature = 0.6.
    \item \passmetric{100} - We use a sampling budget ($N$) = 200 and temperature = 0.8.
\end{enumerate}

\paragraph{Finetuning details}
We finetune the \cllamaB{7} model using deepspeed huggingface trainer.
We use the following training configuration for our main experiments -

\begin{table}[!h]
    \centering
    \begin{tabular}{|l|r|}
        \hline
        \textbf{Training Parameters} & \textbf{Values}                  \\
        \hline
        LR                           & $5e^{-5}$                        \\
        Epochs                       & 1 or 2 depending on the dataset  \\
        Batch Size                   & 256 (combing grad. accumulation) \\
        Dtype                        & bf16                             \\
        \hline
    \end{tabular}
    \label{tab:my_label}
\end{table}
%\subse

\newpage
\section{Code Transformations Implementation}
\label{sec:app:transformimpl}
We implement our code transformation approach using zero-shot prompting with \gptturbo{} model. After transformation, we extract the generated code and evaluate its functional correctness using the provided test cases. In case the program does not pass, we retry the process with up to a maximum of 5 attempts. In our experience, instruction-tuned models can follow precise commands and transform programs very well.

\subsection{Renaming}
We use the following prompt to perform renaming.

\begin{mdframed}[leftmargin=0pt,rightmargin=0pt,innerrightmargin=0pt,innerleftmargin=4pt,innertopmargin=0.9pt,innerbottommargin=0.5pt,roundcorner=5pt,backgroundcolor=backcolour]
\begin{lstlisting}[style=planstyle, basicstyle=\scriptsize]

QUESTION: 
{problem_statement}

ANSWER:
```python
{solution}
```
Rename the variables in the program to be descriptive, meaningful, and consistent. Do not change the original semantics of the program. Enclose the program within backticks as shown above and remember to use descriptive variable names.

\end{lstlisting}
\end{mdframed}

\subsection{Modularization}
Unlike renaming, we perform two rounds of modularization in case the generated program consists of long function implementations (hinting that the function can be decomposed further). We use the following prompt to perform the first round of modularization

\begin{mdframed}[leftmargin=0pt,rightmargin=0pt,innerrightmargin=0pt,innerleftmargin=4pt,innertopmargin=0.9pt,innerbottommargin=0.5pt,roundcorner=5pt,backgroundcolor=backcolour]
\begin{lstlisting}[style=planstyle, basicstyle=\scriptsize]

QUESTION: 
{problem_statement}

ANSWER:
```python
{renamed_solution}
```
Refactor the above program. Follow the guidelines
* make the program more modular with smaller and meaningful helper functions
* good descriptive names for the helper functions
* have an entry function called `main()`
* `main()` is called inside `if __name__ == '__main__'`

Do not change the original semantics of the program significantly and no need to perform optimizations. Enclose the program within backticks as shown above
\end{lstlisting}
\end{mdframed}

Next, in case the modularized program contains a function with the number of lines greater than 20, we further prompt the model while signaling which functions to further decompose. This occurs in about 20-40\% of modularized solutions and we use the following prompt.

\begin{mdframed}[leftmargin=0pt,rightmargin=0pt,innerrightmargin=0pt,innerleftmargin=4pt,innertopmargin=0.9pt,innerbottommargin=0.5pt,roundcorner=5pt,backgroundcolor=backcolour]
\begin{lstlisting}[style=planstyle, basicstyle=\scriptsize]

QUESTION: 
{problem_statement}

ANSWER:
```python
{modularized_solution}
```
Refactor the above program by modularizing it and breaking down long and complex functions into smaller meaningful helper functions. Particularly refactor and decompose the following function(s) into smaller helper functions - {function_names_string}
Only return the refactored program enclosed in backticks as shown above."""
\end{lstlisting}
\end{mdframed}

\subsection{Planning}
We use the following prompt to generate natural language plans

\begin{mdframed}[leftmargin=0pt,rightmargin=0pt,innerrightmargin=0pt,innerleftmargin=4pt,innertopmargin=0.9pt,innerbottommargin=0.5pt,roundcorner=5pt,backgroundcolor=backcolour]
\begin{lstlisting}[style=planstyle, basicstyle=\scriptsize]

QUESTION: 
{problem_statement}

ANSWER:
```python
{modularized_solution}
```
Generate a summary for the following functions and classes in the program within four lines each. The summaries should be descriptive and helpful for understanding the program (however yet concise in four lines). 
The functions and classes are -
{list_of_function_names}
Follow the provided format for the summaries while being informative and concise. Enclose the signatures in backticks as shown above.
\end{lstlisting}
\end{mdframed}

\newpage
\section{Additional Results}
\label{sec:app:results}
\rebuttal{
\subsection{\gptfour{} Judge Evaluation for the transformations}
\label{app:sec:gpt4_eval}
We here present some quantitative evidence of the improvements made from our transformation approach. However, since the transformations are free-form code generation, we rely on using \gptfour{} \emph{as judge}, an evaluation approach gaining popularity for evaluating free-form language outputs~\citep{zheng2023judging,zhuo2023large}. Specifically, we ask the language model to answer whether the modularized refactored code has better variable names, better function decomposition, and is consistent with the original program. The model can provide answers on a key of 1-3 from comparison questions and 0-1 for the consistency question. 
The following prompt depicts our approach
}
\begin{mdframed}[leftmargin=0pt,rightmargin=0pt,innerrightmargin=0pt,innerleftmargin=4pt,innertopmargin=0.9pt,innerbottommargin=0.5pt,roundcorner=5pt,backgroundcolor=backcolour]
\begin{lstlisting}[style=planstyle, basicstyle=\scriptsize]
SYSTEM PROMPT: 
Please act as an impartial judge and evaluate the code refactoring below. You need to evaluate whether the refactored program uses better and correct variable names, refactors the implementation into correct smaller helper functions and consistency with the original program. Your evaluation should be based on correctnes and helpfulness of the refactoring in better understanding the problem and also if it is still consistent with the original program, i.e. it follows similar program logic and algorithm. 

* For evaluating variable names and function decomposition, please give a score from 1 to 3 where 1 means the refactoring is not helpful at all, 2 means the refactoring is somewhat helpful and 3 means the refactoring is very helpful. Example format

Variable names reasoning: [[reasoning for the variable names score, often assessing whether the variable names are more descriptive and meaningful and correctly reflect the variable's purpose]]
Variable names: [[1]] or [[2]] or [[3]]

Function decomposition reasoning: [[reasoning for the decomposition score, often assessing whether some function is too long, possibility to perform further abstractions, choice of abstractions, helper function names]]
Function decomposition: [[1]] or [[2]] or [[3]]

* For evaluating consistency, please give a score of 0 if the refactored program is not consistent with the original program and 1 if it is consistent. Example format

Consistency reasoning: [[reasoning for the consistency score, often assessing whether the refactored program follows similar program logic and algorithm as the original program]]
Consistency: [[0]] or [[1]]

QUESTION: 
{problem_statement}

ORIGINAL SOLUTION:
{solution}

REFACTORED SOLUTION:
{solution}

\end{lstlisting}
\end{mdframed}

\rebuttal{
While this evaluation might portray certain subtle biases, we believe it still provides us a signal to assess the quality of the transformations.
To reduce costs, we the \gptfour{} as judge evaluation to 1000 problems in the \apps{} dataset \footnote{Our prompt spans around 1.5-2k tokens including problem, original, and refactored programs leading to high costs}. 
GPT-4 followed the proposed format for 998 solutions and we present results on them in Table~\ref{app:tab:gpt4_judge}. Results demonstrate that most of the applied transformations are meaningful while remaining consistent with the original ground truth solutions. 
}

\begin{table}[!h]
    \centering
    \begin{tabular}{c|c|c}
            \firsthline
         & Score distribution & Average \\
    \hline

        Variable names & $\{3: 967, 2: 28, 1: 3\}$ & 2.96 \\
        Function decomposition & $\{{3: 938, 2: 59, 1: 1}\}$ & 2.93\\
        Consistency & $\{1: 994, 0: 4\}$ & 0.994 \\
    \hline
    \end{tabular}
    \caption{\textbf{\gptfour{} as a judge evaluation for quality of 998 transformed examples}. We compared the original and unmodified solution using \gptfour{} for variable names used, function decomposition, and consistency of the modified and original solution. Results demonstrate that most of the transformations are successful and meaningful while being consistent with the original program.}
    \label{app:tab:gpt4_judge}
\end{table}

\rebuttal{
To get better insights into \gptfour{} evaluations, we look at the examples which receive lower scores. The scores and associated reasoning appear meaningful. For example, in Figure~\ref{fig:appendix:example-11}, the modularized program is already significantly more readable than original renamed program. However, \gptfour{} identifies that the \code{calculate_permutation_even} and \code{calculate_permutation_even} helper functions are virtually the same and can be abstracted further. Note that this transformation is an artifiact of the fact that original program consisted of same program logic distributed across two far apart if conditions. Similarly, in Figure~\ref{fig:appendix:example-12}, \gptfour{} identified some unmodified variable names like \code{t} while acknowledging other improvements such as \code{sky} to \code{heights} giving it a rating of 2. The rating of 1 is provided when the transformation does not modify any variable names or does not decompose existing functions, as evidenced by score distributed, a rare occurrence. Indeed, often the examples  marked with a rating 2 actually improve upon orignal code in non-trivial ways.\footnote{Curiously, \gptfour{} sometimes returned a rating of 2.5 instead of integer 2 or 3. We rounded it to 2, thus making our evaluation harsher!}
}

\subsection{\appscompete{} Results}
We present the results on \appscompete{} dataset here.

\begin{table}[!h]
    \begin{center}

        \begin{tabular}[h]{lrrr}
            %\hline
            \firsthline
                                 &                                      &                         &                          \\[-0.35cm]
                                 & \multicolumn{3}{c}{\appscompete{}{}}                                                      \\[0.03cm]
            \cline{2-4}
                                 &                                      &                         &                          \\[-0.35cm]
                                 & \passmetric{1}                       & \passmetric{10}         & \passmetric{100}         \\[0.03cm]

            \hline
                                 &                                      &                         &                          \\[-0.35cm]
            \textbf{Fine-tuning} &                                      &                         &                          \\[0.03cm]
            \hline
                                 &                                      &                         &                          \\[-0.35cm]
            \cl{} + \origD{}     & 0.2                                  & 1.7                     & 3.1                      \\[0.03cm]
            \cl{} + \modularD{}  & 0.5                                  & 2.3                     & 3.2                      \\
            [0.03cm]
                                 & \greenbg{\textbf{+0.3}}              & \greenbg{\textbf{+0.6}} & \yellowbg{\textbf{+0.1}} \\[0.03cm]
            \codedavinci{}       & 0.3                                  & 2.9                     & 5.7                      \\
            \hline
        \end{tabular}
        \caption{\textbf{Results on the \appscompete{} dataset.}}
        \label{tab:appscomp}
    \end{center}
\end{table}

\rebuttal{
    \subsection{Ablation on choice of model}
    \label{subsec:app:model_ablation}
    We use \gptturbo{} as the default model for performing the transformations in the main experiments since it provides a nice balance between the accuracy and cost of performing the transformations. Here, to demonstrate the generality of our approach we perform an ablation by replacing the transformation model with \gptfourturbo{}. Since this model is about 8-10x more expensive than \gptturbo{}, we perform this ablation on a subset of 5k programs sampled from the dataset.

    \textbf{Experimental setup.} We repeat the renaming and modularization steps described in Section~\ref{subsec:setup:lmdt} using the \gptfourturbo{} model. We call the resulting transformed dataset as \modularDfour{}. Next, fairly compare the resulting dataset with the original and modularized dataset generated using \gptturbo{}, we sample the corresponding parallel original and transformed programs and call them \origD{} and \modularDthreefive{} datasets.

    \begin{table}[!h]
    \begin{center}

        \begin{tabular}[h]{lrrr}
            %\hline
            \firsthline
                                         &                                  &                         &                         \\[-0.35cm]
                                         & \multicolumn{3}{c}{\appsintro{}}                                                     \\[0.03cm]
            \cline{2-4}
                                         &                                  &                         &                         \\[-0.35cm]
                                         & \passmetric{1}                   & \passmetric{10}         & \passmetric{100}        \\[0.03cm]

            \hline
                                         &                                  &                         &                         \\[-0.35cm]
            \cl{} + \origD{}             & 16.3                             & 31.6                    & 37.6                    \\[0.03cm]
            \cl{} + \modularDthreefive{} & 18.8                             & 33.0                    & 38.2                    \\
            [0.03cm]
            \cl{} + \modularDfour{}      & 19.4                             & 34.3                    & 40.0                    \\
            [0.03cm]
                                         & \greenbg{\textbf{+0.6}}          & \greenbg{\textbf{+1.3}} & \greenbg{\textbf{+1.8}} \\[0.03cm]
            \hline
        \end{tabular}
        \caption{\textbf{Ablation on the choice of model used for performing the transformations.} \modularDthreefive{} represents the dataset generated using \gptturbo{} and \modularDfour{} represents the dataset generated using \gptfourturbo{}. We find that the performance of the model trained on the \modularDfour{} dataset is better than the model trained on the \modularDthreefive{} dataset.}
        \label{tab:app:model_ablation}
    \end{center}
\end{table}

}

\newpage
\section{Additional Related Work}
\label{sec:app:related}
\rebuttal{
Code \llms{} have been used for multiple domains in various lines of approaches. Here, we present a few key approaches and recommend the reader to ~\cite{hou2023large} for a detailed survey. 
~\cite{chen2021evaluating} released the \codedavinci{} model and evaluate it for code generation. 
Since then, \llms{} have been used for a variety of domains such as 
data science~\citep{Jigsaw,Lai2022DS1000,yin-etal-2023-natural}, 
APIs~\citep{wang2022odex,patil2023gorilla}, and 
repositories~\citep{zhang2023repocoder, bairi2023codeplan,shrivastava2023repository}.
~\citep{le2022coderl,shojaee2023executionbased,liu2023rltf} use reinforcement learning with compilation/execution feedback to fine-tune code \llms{} for (algorithmic) code generation task. 

Other works have approached code generation from different fronts, exploring planning~\citep{jiang2023self}, repair~\citep{chen2023teaching,shinn2023reflexion}, test generation~\citep{key2022speak,chen2022codet}, and prompt optimization~\citep{liu2023improving}.
}

\newpage
\section{Examples of Transformed Program}
\label{sec:app:examples}
\begin{figure}[!h]
    \centering
    \begin{subfigure}{0.48\textwidth}
        \begin{mdframed}[leftmargin=0pt,rightmargin=0pt,innerrightmargin=3pt,innerleftmargin=3pt,innertopmargin=0.5pt,innerbottommargin=0.5pt,roundcorner=5pt,backgroundcolor=backcolour]
            \input{figures/appendix/examples/example-1/original}
        \end{mdframed}
        \caption{Original program}
    \end{subfigure}\hfil%
    \begin{subfigure}{0.48\textwidth}
        \begin{mdframed}[leftmargin=0pt,rightmargin=0pt,innerrightmargin=3pt,innerleftmargin=3pt,innertopmargin=0.5pt,innerbottommargin=0.5pt,roundcorner=5pt,backgroundcolor=backcolour]
            \input{figures/appendix/examples/example-1/modularized}
        \end{mdframed}
        \caption{Transformed program}
    \end{subfigure}
    \caption{Original and transformed programs}
    \label{fig:appendix:example-1}
\end{figure}
\newpage
\begin{figure}[!h]
    \centering
    \begin{subfigure}{0.48\textwidth}
        \begin{mdframed}[leftmargin=0pt,rightmargin=0pt,innerrightmargin=3pt,innerleftmargin=3pt,innertopmargin=0.5pt,innerbottommargin=0.5pt,roundcorner=5pt,backgroundcolor=backcolour]
            \input{figures/appendix/examples/example-2/original}
        \end{mdframed}
        \caption{Original program}
    \end{subfigure}\hfil%
    \begin{subfigure}{0.48\textwidth}
        \begin{mdframed}[leftmargin=0pt,rightmargin=0pt,innerrightmargin=3pt,innerleftmargin=3pt,innertopmargin=0.5pt,innerbottommargin=0.5pt,roundcorner=5pt,backgroundcolor=backcolour]
            \input{figures/appendix/examples/example-2/modularized}
        \end{mdframed}
        \caption{Transformed program}
    \end{subfigure}
    \caption{Original and transformed programs}
    \label{fig:appendix:example-2}
\end{figure}
\newpage
\begin{figure}[!h]
    \centering
    \begin{subfigure}{0.48\textwidth}
        \begin{mdframed}[leftmargin=0pt,rightmargin=0pt,innerrightmargin=3pt,innerleftmargin=3pt,innertopmargin=0.5pt,innerbottommargin=0.5pt,roundcorner=5pt,backgroundcolor=backcolour]
            \input{figures/appendix/examples/example-3/original}
        \end{mdframed}
        \caption{Original program}
    \end{subfigure}\hfil%
    \begin{subfigure}{0.48\textwidth}
        \begin{mdframed}[leftmargin=0pt,rightmargin=0pt,innerrightmargin=3pt,innerleftmargin=3pt,innertopmargin=0.5pt,innerbottommargin=0.5pt,roundcorner=5pt,backgroundcolor=backcolour]
            \input{figures/appendix/examples/example-3/modularized}
        \end{mdframed}
        \caption{Transformed program}
    \end{subfigure}
    \caption{Original and transformed programs}
    \label{fig:appendix:example-3}
\end{figure}
\newpage
\begin{figure}[!h]
    \centering
    \begin{subfigure}{0.48\textwidth}
        \begin{mdframed}[leftmargin=0pt,rightmargin=0pt,innerrightmargin=3pt,innerleftmargin=3pt,innertopmargin=0.5pt,innerbottommargin=0.5pt,roundcorner=5pt,backgroundcolor=backcolour]
            \input{figures/appendix/examples/example-4/original}
        \end{mdframed}
        \caption{Original program}
    \end{subfigure}\hfil%
    \begin{subfigure}{0.48\textwidth}
        \begin{mdframed}[leftmargin=0pt,rightmargin=0pt,innerrightmargin=3pt,innerleftmargin=3pt,innertopmargin=0.5pt,innerbottommargin=0.5pt,roundcorner=5pt,backgroundcolor=backcolour]
            \input{figures/appendix/examples/example-4/modularized}
        \end{mdframed}
        \caption{Transformed program}
    \end{subfigure}
    \caption{Original and transformed programs}
    \label{fig:appendix:example-4}
\end{figure}
\newpage
\begin{figure}[!h]
    \centering
    \begin{subfigure}{0.48\textwidth}
        \begin{mdframed}[leftmargin=0pt,rightmargin=0pt,innerrightmargin=3pt,innerleftmargin=3pt,innertopmargin=0.5pt,innerbottommargin=0.5pt,roundcorner=5pt,backgroundcolor=backcolour]
            \input{figures/appendix/examples/example-5/original}
        \end{mdframed}
        \caption{Original program}
    \end{subfigure}\hfil%
    \begin{subfigure}{0.48\textwidth}
        \begin{mdframed}[leftmargin=0pt,rightmargin=0pt,innerrightmargin=3pt,innerleftmargin=3pt,innertopmargin=0.5pt,innerbottommargin=0.5pt,roundcorner=5pt,backgroundcolor=backcolour]
            \input{figures/appendix/examples/example-5/modularized}
        \end{mdframed}
        \caption{Transformed program}
    \end{subfigure}
    \caption{Original and transformed programs}
    \label{fig:appendix:example-5}
\end{figure}
\newpage
\begin{figure}[!h]
    \centering
    \begin{subfigure}{0.48\textwidth}
        \begin{mdframed}[leftmargin=0pt,rightmargin=0pt,innerrightmargin=3pt,innerleftmargin=3pt,innertopmargin=0.5pt,innerbottommargin=0.5pt,roundcorner=5pt,backgroundcolor=backcolour]
            \input{figures/appendix/examples/example-6/original}
        \end{mdframed}
        \caption{Original program}
    \end{subfigure}\hfil%
    \begin{subfigure}{0.48\textwidth}
        \begin{mdframed}[leftmargin=0pt,rightmargin=0pt,innerrightmargin=3pt,innerleftmargin=3pt,innertopmargin=0.5pt,innerbottommargin=0.5pt,roundcorner=5pt,backgroundcolor=backcolour]
            \input{figures/appendix/examples/example-6/modularized}
        \end{mdframed}
        \caption{Transformed program}
    \end{subfigure}
    \caption{Original and transformed programs}
    \label{fig:appendix:example-6}
\end{figure}
\newpage
\begin{figure}[!h]
    \centering
    \begin{subfigure}{0.48\textwidth}
        \begin{mdframed}[leftmargin=0pt,rightmargin=0pt,innerrightmargin=3pt,innerleftmargin=3pt,innertopmargin=0.5pt,innerbottommargin=0.5pt,roundcorner=5pt,backgroundcolor=backcolour]
            \input{figures/appendix/examples/example-7/original}
        \end{mdframed}
        \caption{Original program}
    \end{subfigure}\hfil%
    \begin{subfigure}{0.48\textwidth}
        \begin{mdframed}[leftmargin=0pt,rightmargin=0pt,innerrightmargin=3pt,innerleftmargin=3pt,innertopmargin=0.5pt,innerbottommargin=0.5pt,roundcorner=5pt,backgroundcolor=backcolour]
            \input{figures/appendix/examples/example-7/modularized}
        \end{mdframed}
        \caption{Transformed program}
    \end{subfigure}
    \caption{Original and transformed programs}
    \label{fig:appendix:example-7}
\end{figure}
\newpage
\begin{figure}[!h]
    \centering
    \begin{subfigure}{0.48\textwidth}
        \begin{mdframed}[leftmargin=0pt,rightmargin=0pt,innerrightmargin=3pt,innerleftmargin=3pt,innertopmargin=0.5pt,innerbottommargin=0.5pt,roundcorner=5pt,backgroundcolor=backcolour]
            \input{figures/appendix/examples/example-10/original}
        \end{mdframed}
        \caption{Original program}
    \end{subfigure}\hfil%
    \begin{subfigure}{0.48\textwidth}
        \begin{mdframed}[leftmargin=0pt,rightmargin=0pt,innerrightmargin=3pt,innerleftmargin=3pt,innertopmargin=0.5pt,innerbottommargin=0.5pt,roundcorner=5pt,backgroundcolor=backcolour]
            \input{figures/appendix/examples/example-10/modularized}
        \end{mdframed}
        \caption{Transformed program}
    \end{subfigure}
    \caption{Original and transformed programs}
    \label{fig:appendix:example-10}
\end{figure}
\newpage
\begin{figure}[!h]
    \centering
    \begin{subfigure}{0.48\textwidth}
        \begin{mdframed}[leftmargin=0pt,rightmargin=0pt,innerrightmargin=3pt,innerleftmargin=3pt,innertopmargin=0.5pt,innerbottommargin=0.5pt,roundcorner=5pt,backgroundcolor=backcolour]
            \input{figures/appendix/examples/example-9/original}
        \end{mdframed}
        \caption{Original program}
    \end{subfigure}\hfil%
    \begin{subfigure}{0.48\textwidth}
        \begin{mdframed}[leftmargin=0pt,rightmargin=0pt,innerrightmargin=3pt,innerleftmargin=3pt,innertopmargin=0.5pt,innerbottommargin=0.5pt,roundcorner=5pt,backgroundcolor=backcolour]
            \input{figures/appendix/examples/example-9/modularized}
        \end{mdframed}
        \caption{Transformed program}
    \end{subfigure}
    \caption{Original and transformed programs}
    \label{fig:appendix:example-9}
\end{figure}
\newpage
\begin{figure}[!h]
    \centering
    \begin{subfigure}{0.48\textwidth}
        \begin{mdframed}[leftmargin=0pt,rightmargin=0pt,innerrightmargin=3pt,innerleftmargin=3pt,innertopmargin=0.5pt,innerbottommargin=0.5pt,roundcorner=5pt,backgroundcolor=backcolour]
            \input{figures/appendix/examples/example-8/original}
        \end{mdframed}
        \caption{Original program}
    \end{subfigure}\hfil%
    \begin{subfigure}{0.48\textwidth}
        \begin{mdframed}[leftmargin=0pt,rightmargin=0pt,innerrightmargin=3pt,innerleftmargin=3pt,innertopmargin=0.5pt,innerbottommargin=0.5pt,roundcorner=5pt,backgroundcolor=backcolour]
            \input{figures/appendix/examples/example-8/modularized}
        \end{mdframed}
        \caption{Transformed program}
    \end{subfigure}
    \caption{Original and transformed programs}
    \label{fig:appendix:example-8}
\end{figure}

\newpage
\begin{figure}[!h]
    \centering
    \begin{subfigure}{0.48\textwidth}
        \begin{mdframed}[leftmargin=0pt,rightmargin=0pt,innerrightmargin=3pt,innerleftmargin=3pt,innertopmargin=0.5pt,innerbottommargin=0.5pt,roundcorner=5pt,backgroundcolor=backcolour]
            \input{figures/appendix/examples/example-11/original}
        \end{mdframed}
        \caption{Original program}
    \end{subfigure}\hfil%
    \begin{subfigure}{0.48\textwidth}
        \begin{mdframed}[leftmargin=0pt,rightmargin=0pt,innerrightmargin=3pt,innerleftmargin=3pt,innertopmargin=0.5pt,innerbottommargin=0.5pt,roundcorner=5pt,backgroundcolor=backcolour]
            \input{figures/appendix/examples/example-11/modularized}
        \end{mdframed}
        \caption{Transformed program}
    \end{subfigure}
    \caption{Original and transformed programs}
    \label{fig:appendix:example-11}
\end{figure}

\newpage
\begin{figure}[!h]
    \centering
    \begin{subfigure}{0.48\textwidth}
        \begin{mdframed}[leftmargin=0pt,rightmargin=0pt,innerrightmargin=3pt,innerleftmargin=3pt,innertopmargin=0.5pt,innerbottommargin=0.5pt,roundcorner=5pt,backgroundcolor=backcolour]
            \input{figures/appendix/examples/example-12/original}
        \end{mdframed}
        \caption{Original program}
    \end{subfigure}\hfil%
    \begin{subfigure}{0.48\textwidth}
        \begin{mdframed}[leftmargin=0pt,rightmargin=0pt,innerrightmargin=3pt,innerleftmargin=3pt,innertopmargin=0.5pt,innerbottommargin=0.5pt,roundcorner=5pt,backgroundcolor=backcolour]
            \input{figures/appendix/examples/example-12/modularized}
        \end{mdframed}
        \caption{Transformed program}
    \end{subfigure}
    \caption{Original and transformed programs}
    \label{fig:appendix:example-12}
\end{figure}

\end{document}